# Multi-modal Egocentric Activity Recognition using Audio-Visual Features

Mehmet Ali Arabacı, Fatih Özkan, Elif Surer, Peter Jančovič, Alptekin Temizel

***Abstract*—Egocentric activity recognition in first-person videos has an increasing importance with a variety of applications such as lifelogging, summarization, assisted-living and activity tracking. Existing methods for this task are based on interpretation of various sensor information using pre-determined weights for each feature. In this work, we propose a new framework for egocentric activity recognition problem based on combining audio-visual features with multi-kernel learning (MKL) and multi-kernel boosting (MKBoost). For that purpose, firstly grid optical-flow, virtual-inertia feature, log-covariance, cuboid are extracted from the video. The audio signal is characterized using a 'supervector', obtained based on Gaussian mixture modelling of frame-level features, followed by a maximum a-posteriori adaptation. Then, the extracted multi-modal features are adaptively fused by MKL classifiers in which both the feature and kernel selection/weighing and recognition tasks are performed together. The proposed framework was evaluated on a number of egocentric datasets. The results showed that using multi-modal features with MKL outperforms the existing methods.**

## I. Introduction

INCREASING use of social media has led to a huge amount of multimedia content sharing where videos are as important as text and audio-based posts. The main driving factor behind the increase in video is the widespread use of handheld cameras, sports and action cameras (i.e. GoPro), mobile phones and wearable cameras as accessories (i.e. Snap Spectacles, Google Glass). For this reason, first-person (i.e. egocentric) videos, in which the actors are getting involved in the activities or events, have become popular. In first-person videos, the world is seen from the perspective of the actor within the context of the actor's activities and goals. Some of the application areas of egocentric video analysis are: activity recognition, egocentric video summarization, attention localization and daily care of elderly and disabled patients. Most of these fields are still being studied as research problems given the unstructured nature and motion dynamics of first-person videos.

Popular use of egocentric videos in daily life has created the need of summarizing, organizing and analyzing them. A recent survey on summarization of egocentric videos reveals the diversity and abundance of the objectives, approaches and evaluation strategies [1]. Current activity recognition algorithms are mostly focused on analysis of third-person videos [2, 3, 4, 5].

On the other hand, egocentric videos have different characteristics than third-person videos (i.e. fast scene transitions, rapid change in illumination, motion blur and high ego-motion) and require particular approaches.

The studies on egocentric activity recognition can be grouped as object and motion-based approaches [6]. In object-based methods, activity recognition is performed using the object(s) detected in videos (i.e. detection of cheese and bread objects imply "making cheese sandwich" activity) [7]. The performance of object-based approaches is directly dependent on the performance of object recognition methods and they are vulnerable to occlusions. Motion-based approaches make use of the assumption that different types of activities such as running, walking, stair climbing, and writing involve different body motions, and these motion patterns can be used for recognizing activities [8, 9]. To obtain this motion pattern, various types of sensors are used such as inertial measurement units (IMUs), GPS or eye-trackers with more specialized equipment (such as SenseCam, Google Glass) in addition to optic and audio sensors. However, these special devices are not widely used as they are limited in battery life and storage capacities.

The methods for egocentric activity recognition use various types of features and present different strategies to learn the activities such as supervised or unsupervised learning. One of the most prominent approaches is to combine multiple features using the data generated by different sensors. In this method, each feature extracted locally or globally is considered as a separate channel having equal weights [10]. While finding the optimal feature combination still remains an open research question, deep learning architectures have been recently investigated for feature selection. Song et al. [11] used a deep neural network architecture for kernel fusion. Sudhakaran and Lanz [12] used Convolutional Neural Networks (CNNs) for frame level feature extraction and their results outperform the state-of-the-art. Another application of deep learning is by Bambach et al. [13] where first-person videos from different people are acquired and their interactions with the objects are trained and modeled based on each person's viewpoint.

This work was partly supported by The Scientific and Technological Research Council of Turkey under TUBITAK BIDEB-2219 grant no 1059B191500048.

Mehmet Ali Arabacı, Fatih Özkan, Elif Surer and Alptekin Temizel are with Graduate School of Informatics, Middle East Technical University (METU), 06800, Ankara, Turkey (e-mail: {mehmet.arabaci, fatih.ozkan, elifs, atemizel}@metu.edu.tr).

Peter Jančovič is with Electronic, Electrical and Systems Engineering, University of Birmingham, Edgbaston, Birmingham, B15 2TT, UK (e-mail: p.jancovic@bham.ac.uk). Alptekin Temizel was with Electronic, Electrical and Systems Engineering, University of Birmingham on sabbatical leave from METU partly during this work.

Feichtenhofer et al. [14] use CNNs for spatio-temporal fusion of video snippets to increase performance boosting.

Support vector machines (SVMs) is a popular kernel-based technique. However, it does not provide a mechanism for effective use of multiple features since each feature might have varying importance and they may require different kernels. Multi-kernel learning approach (MKL) has been proposed for feature fusion using kernel-based classifiers to improve the classification performance [15]. Recently, MKL has been shown to be a promising method since its activity recognition performance results outperform classical approaches [16]. In this data-driven approach, multiple features are fused in an adaptive way using different types of kernels. Even if the base kernels cannot perform well for all features, their parameters and weights are optimized to get the best performance by using complementary information coming from different types of features. By this way, features are dynamically weighted at the training stage that allows to create adaptive solutions for different first-person activity recognition problems.

In this work, we use audio-visual information that can be acquired by off-the-shelf cameras without any need for special equipment or sensors. The features are extracted using different types of information such as optical flow, intensity gradient, video-based inertia and audio. On the other hand, using multi-modality introduces the problem of feature selection and decision fusion. At that point, MKL method was preferred which performs both the feature selection and recognition tasks concurrently [17] that also allows easy integration of any additional sensor to the proposed framework. Additionally, adaptive weighting of features according to their classification performances on base classifiers makes the framework robust against non-informative data.

## II. Proposed Framework

The proposed framework is shown in Fig. 1. Both audio and video information were used to model different activities. Then, MKL is used to learn the weights of different features, kernels and their parameters using training videos. Finally, egocentric activity recognition is performed for test videos considering previously selected features, base kernels and their parameters. Details of these steps are given in the following sections.

### *A. Feature Extraction*

The features used in this work are global video features, local video features and audio features that are explained in the following sections in detail. Our motivation is that the use of complementary features from different modalities would contribute to the recognition performance of egocentric activities.

#### *1) Global Video Features*

Global video features are extracted using the whole video frame and they do not contain any local motion information. *Grid Optical Flow-Based Features (GOFF)* are motion-based video features extracted from spatio-temporal information specifically designed for first-person videos [9]. *Vision-based inertial features* (*VIF)* [9] are used to approximate inertia data (velocity and acceleration) to model egocentric activities. Log-Covariance (*Log-C)* [18] features are dense video features derived from the optical flow data as well as intensity gradient. The following sections explain these global features in detail.

##### *a) GOFF*

*GOFF* is used to model the discriminative motion patterns in the optical flow data such as magnitude, direction and frequency that was originally proposed by [9]. For that purpose, Motion Magnitude Histogram Features (*MMHF*), Motion Direction Histogram Features (*MDHF*), Motion Direction Histogram Standard-Deviation Feature (*MDHSF*), Fourier Transform of Motion Direction Access Frame (*FTMAF*) and Fourier Transform of Grid Motion Per-Frame (*FTMPF*) features are extracted from video frames divided into grids. Motion estimation is performed by Farneback Optical Flow algorithm [19].

*MMHF* is the histogram representation of grid optical flow magnitude values in which a non-uniform quantization process with 15 levels is used [9]. *MDHF* is another histogram representation of grid optical flow considering its quantized direction values. *MDHF* was uniformly quantized into 36 levels that corresponds to $10^o$ between each level. *MDHSF* represents a 36 dimensional vector that includes the standard deviation of each direction bin across the temporal dimension. *FTMAF* is a frequency-based feature that measures the variation for each direction bin along temporal dimension using decomposed frequency bands. In contrast to *MDHSF*, *FTMAF* quantifies the detailed dynamics of motion direction into 25 levels. Lastly, *FTMPF* measures the variation of grid optical flow within a frame that also has 25 levels. As a result, *GOFF* has 137 dimensions after concatenating all of the sub-features.

##### *b) VIF*

*VIF* provides virtual inertial information derived by using intensity centroid across frames in a video without physically using inertial sensors [9]. Three different sub-features extracted in temporal dimension were used for that purpose: *zero-crossing (ZC)*, *4MEKS* and frequency-domain feature (FF). *ZC* uses velocity and acceleration values generated from intensity centroid for each frame and measures zero-crossing rates of velocity and acceleration values. *4MEKS* represents the time-domain features in which minimum, maximum, median, energy, kurtosis, mean and standard deviation values are calculated for each inertial signal. *FF* feature holds low frequency components of the variations in velocity and acceleration. In this work, the number of frequency components was selected as 10. Similar to *GOFF*, all sub-features of VIF were concatenated that makes the resulting feature a 106 dimensional vector.

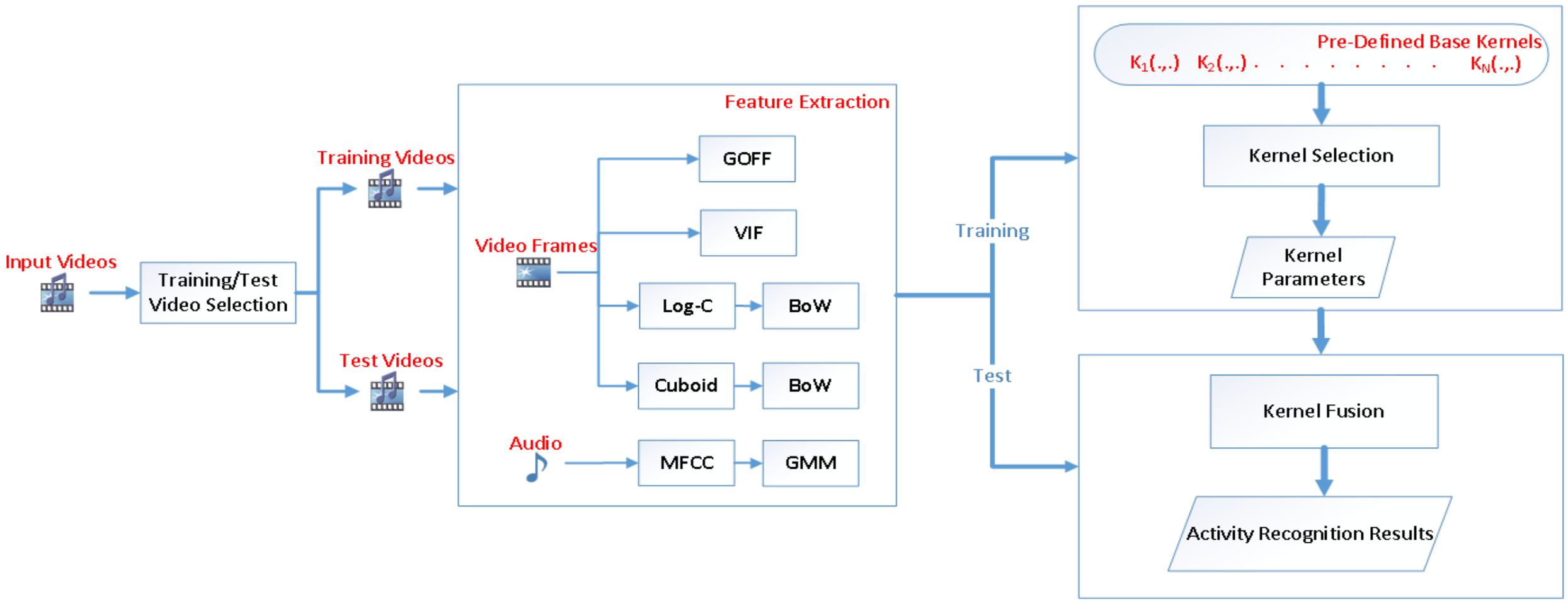

Fig. 1. The proposed solution for egocentric activity recognition using audio-visual features.

*c) Log-C*

Feature covariance matrix is an effective way of representing dense set of localized features. Bag of local features can be represented in a lower dimension by the help of feature covariance matrices. In this work, feature covariance matrix was determined by using optical flow and gradient vectors. For each pixel of a video frame, a 12x12 dimensional covariance matrix was calculated by using intensity gradient of raw video sequences with respect to temporal direction and first-order partial derivative of optical flow with respect to spatial x and y directions, spatial divergence, vorticity, gradient tensor and rate of strain tensor [18].

The dimension of the covariance matrix is only related to the dimension of the feature vectors (i.e. 12x12 in this work). Covariance matrices lie on the Riemannian manifold and we use matrix logarithm [20] operation to convert manifold of covariance matrices into Euclidean. As a result, the feature vector size was reduced to 78 due to its symmetry. After that, the extracted feature vector was normalized by standard deviation and clustered using k-means for each video segment. Finally, a descriptor is defined using Bag-of-Words (*BoW*) for each single activity video. *BoW* size was set to 300. However, principal component analysis (*PCA*) was applied to the descriptor in order to reduce the dimension of sparse *BoW* vectors except the classifier that uses histogram intersection kernels.

*2) Local Video Features*

*Cuboids* have been used successfully for activity recognition problem before [4]. In addition to global video features, a sparse 3D XYT space-time feature *cuboid* [21] is used as a local video feature.

*Cuboid* feature was developed as an alternative to 2D interest point detectors which considers temporal dimensions in addition to the spatial dimensions. Before the feature extraction process, interest point detector was employed to detect the corners in spatio-temporal dimensions by responding strongly to the local areas containing motion and including spatio-temporal corners. After that, a *cuboid* feature was extracted at each interest point that includes brightness gradient and optical flow information [21]. Similar to *Log-C* feature, *cuboid* was also configured to generate descriptors by using *BoW*. The size of the histogram was set as 500 and *PCA* was applied in order reduce the dimension. Histogram intersection kernel was used for the raw histogram feature without applying *PCA*.

*3) Audio Features*

In order to fuse video and audio modalities using the SVM- and MKL-based frameworks, an utterance of audio recording of an activity needs to be mapped into a vector space. To do so, we employed a commonly used methodology in the field of speaker recognition from speech [22, 23].

The first step is to split an audio signal into a sequence of frames and represent each frame using a spectrum-based feature vector. We used Mel-frequency cepstral coefficients (*MFCCs*) [24] as frame features. The discrete Fourier transform was applied on each frame and the resulting magnitude spectrum was passed to a bank of Mel-spaced triangular filters and then discrete cosine transform was applied, providing MFCCs. In the above, we explored a range of values for the parameters, with final setup: the frame length of 40 ms, shift between adjacent frames of 10 ms and 23 filter-bank channels. Only the first 12 *MFCCs* were used. The frame energy was added as the $13^{th}$ feature. These features were appended with their temporal derivatives, referred to as delta and delta-delta coefficients, calculated as in [25], using the span of ±3 and ±2 frames, respectively. This resulted in 39 dimensional feature representation of each signal frame.

We then modeled the distribution of these feature vectors using the Gaussian mixture model (GMM), with diagonal covariance matrices. First, a class-independent model, referred to as the Universal Background Model (*UBM*) was estimated using all of the training data from all classes. A class-dependent *GMM* was then obtained by performing maximum a-posteriori adaptation [26] of the component mean vectors of the *UBM*, using class-specific training data. The mean vectors of the components of the resulting class-dependent *GMM* are then concatenated to form a ‘supervector’ [27]. A supervector is obtained for each utterance of each class, resulting in a set of supervectors per class. Supervectors are then used as a vector representation of each class for activity classification. We performed experiments using different number of GMM components, with similar performance being achieved when using from 16 to 64 mixtures. The use of such small number of mixture components may be due to the small amount of training data available. The reported results are using 16 mixtures. We also explored dimensionality reduction of the supervectors using the PCA but only little effect on the performance was observed.

### B. *Classification*

For classification, in addition to support vector machines (*SVM*), we also used multi-kernel learning (*MKL*) and multi kernel boosting (*MKBoost*) for fusion of different features and selection of the kernels during the training. Additionally, each video segment containing one activity was represented with a single feature vector.

#### 1) *Support Vector Machines and Kernel Selection*

SVM is a kernel-based method and use of kernels allows operating in higher dimensional feature spaces than the original feature space. In this work, we use *SVM* for activity classification using feature vectors that are formed by concatenating all features. The most widely used kernel types for *SVMs* are polynomial and radial basis functions (*RBFs*). In this work, both of them were tested and polynomial kernel was selected since it performed better in comparison to *RBFs*. Additionally, the best kernel parameters were also searched and the order of polynomial kernel was set as 3. The polynomial kernel in the order of $p$ is defined as:

$$\kappa(x_i, x_j) = (\langle x_i, x_j \rangle + l)^p \quad (1)$$

where κ represents kernel function, $x$'s are the features, $p$ is the maximal order of monomials making up the new feature space and $l$ is a bias towards lower order monomial. The intuition behind this kernel definition is that it is often useful to construct new features as products of original features [28].

*Log-C* and *cuboid* features can be described as histogram data because of the *BoW* model. Therefore, we use histogram intersection kernels since they are more appropriate for histogram comparison. In this work, a modified histogram intersection kernel (*DC-Int*) [29] was selected, which is defined as follows:

$$\kappa(x_i, x_j) = exp\left(-\sum_{n=1}^{N} D_c(H_i^n, H_j^n)\right) \quad (2)$$

where $H_i^n$ is a $w$ dimensional histogram of $n^{th}$ channel for $i^{th}$ video and $D_c(H_i^n, H_j^n)$ is the histogram distance defined as:

$$D_n(H_i^n, H_j^n) = 1 - \frac{\sum_{m=1}^{w} min(h_{im}, h_{jm})}{\sum_{m=1}^{w} max(h_{im}, h_{jm})} \quad (3)$$

where $h_{im}$ is the $m^{th}$ visual word identified for $i^{th}$ video.

#### 2) *MKL*

Kernel methods such as *SVM* have proved to be efficient tools for solving classification and regression problems [17]. Data representation is implicitly chosen through the used kernel $\kappa(x, x^i)$. Solution of learning problems for kernels is of the form:

$$\sum_{i=1}^{l} \alpha_i^* \, \kappa\,(x, x_i) + b^* \quad (4)$$

where $\alpha_i^*$ and $b^*$ are coefficients to be learned from examples, while $\kappa(.,.)$ is a given positive definite kernel associated with a reproducing kernel Hilbert space (*RKHS*). It was shown that using multiple kernels ($\kappa_m$) instead of a single one can enhance the interpretability of the decision function and improve performance [30].

$$\kappa\,(x, x^i) = \sum_{m=1}^{M} d_m \, \kappa_m(x, x^i) \quad (5)$$

where $M$ is the number of kernels, $d_m \geq 0$ and $\sum_{m=1}^{M} d_m = 1$. MKL allows learning both the coefficients $\alpha_i$ and the weights $d_m$ in a single optimization. In this work, we used two MKL algorithms that are going to be introduced in the following sections.

##### a) *SimpleMKL*

SimpleMKL offers a solution to *MKL* by using a weighted $l2$-norm normalization. The proposed solution is based on a gradient descent wrapping standard SVM solver that determines the combination of kernels [31]. In this work, SimpleMKL algorithm was used as proposed in [31] without any modification and its pseudo-code is given in Algorithm 1. In this algorithm, $J$ represents the differentiable objective function and $\nabla_D$ shows the gradient descent directions for each step.

**Algorithm 1: SimpleMKL**

INPUT:
set equal kernel weights: $d_m = \frac{1}{M}, i = 1, \ldots, M$
***while*** stopping criterion not met ***do***
compute $J(d)$ by using an SVM solver with $K = \sum_m d_m K_m$
compute $\frac{\partial J}{\partial d_m}$ for $m = 1, \ldots, M$ and descent direction $\nabla_D$
set $\mu = argmax(d_m), J^\dagger = 0, d^\dagger = 0, \nabla_D^\dagger = \nabla_D$
***while*** $J^\dagger < J(d)$ ***do*** {descent direction update}
$d = d^\dagger, \nabla_D = \nabla_D^\dagger$
$v = \underset{\{m \mid \nabla_D^m < 0\}}{argmin} \left(-\frac{d_m}{\nabla_D^m}\right), \gamma_{max} = -\frac{d_v}{\nabla_D^v}$
$d^\dagger = d + \gamma_{max}\nabla_D, \nabla_D^{\mu\dagger} = \nabla_D^\mu - \nabla_D^v, \nabla_D^{v\dagger} = 0$
Compute $J^\dagger$ by using an SVM solver with $K = \sum_m d_m^\dagger K_m$
***end while***
linear search along $D$ for $\gamma \in [0, \gamma_{max}]$
{calls an SVM solver for each trial value}
$d \leftarrow d + \gamma\nabla_D$
***end while***

##### b) *Multiple-Kernel Boosting (MKBoost)*

MKBoost is employed as the boosting framework in order to learn an ensemble of multiple base kernel classifiers, each of which is learned from a single kernel. The combination weights for both the kernels and classifiers can be efficiently determined through the learning process of boosting [32] using a similar procedure to Adaboost [33]. In this approach, some kernel classifiers with multiple kernels $K$ through a series of boosting trials $t = 1, \dots, T$, where $T$ denotes the total number of boosting trials, are repeatedly learned.

At each boosting trial, a distribution of weights $S_t$ is engaged to indicate the importance of the training examples for learning. At each trial, the weights of the wrongly classified examples are increased while the weights of those correctly classified examples are decreased in order to focus on those examples that are hard to be correctly classified. The pseudo-code for the selected MKBoost is given in Algorithm 2.

**Algorithm 2: MKBoost**

1: INPUT:
training data: $(x_1, y_1), \dots, (x_N, y_N)$, labels: $y = \{1, \dots, C\}$
$C$ is the number of class
kernel functions: $K_m(.,.) : X \times X \to \mathbb{R}, j = 1, \dots, M$
initial distribution: $\chi \sim U(0, N)$
2: $\boldsymbol{for\ t = 1, \dots, T\ do}$
3: sample $r * N$ examples $(S_t)$ using distribution $\chi$
4: $\boldsymbol{for\ j = 1, \dots, M\ do}$
5: train weak classifier with kernel $K_j$
$K_j^t : S_t \to \{1, \dots, C\}$
6: compute the training error over $S_t$
$\in_t^j = \sum_{i=1}^{N} S_t(i) f_t^i(x_i) \neq y_i$
7: $\boldsymbol{end\ for}$
8: select the best classifier with the minimum error rate

$$f_t = arg \min_{f_t^j} \in_t^j = arg \min_{f_t^j} \sum_{i=1}^{N} S_t(i)\left(K_t^j(x_i) \neq y_i\right)$$

9: choose $w_t = \frac{1}{2} ln\left(\frac{1-\epsilon_t}{\epsilon_t}\right)$, where $\in_t = \min_{j \in \{1, \dots, M\}} \in_t^j$
10: update $S_{t+1}(i)$:

$$S_{t+1}(i) = \frac{S_t(i)}{Z_t} x \{ \begin{matrix} e^{-w_t} & if\ K_t(x_i) = y_i \\ e^{w_t} & if\ K_t(x_i) \neq y_i \end{matrix}$$

$Z_t$ is a normalization factor to make $S_t$ a distribution
11: $\boldsymbol{end\ for}$
12: OUTPUT: $\boldsymbol{f(x) = sign(\sum_{t=1}^{T} w_t K_x(t))}$

## III. Experimental Results

The performance of the proposed framework was evaluated using three egocentric video datasets: JPL [10], MEAD [34], and DogC [29]. Global (*GOFF*, *VIF*, *Log-C*) and local (*cuboid*) video features were extracted for all the datasets. Audio feature was extracted only for MEAD because it is the only one having audio information.

JPL First-Person Interaction dataset [10] is composed of first-person videos of interaction-level activities by 8 different actors. It contains 4 positive (i.e. friendly) interactions with the observer (*shaking hand*, *hugging*, *pet*, *waving hand*), 1 neutral interaction (*pointing*), and 2 negative (i.e. hostile) interactions (*punching*, *throwing objects*) for each actor. There are a total of 84 videos at 320x240 at 30fps.

The Multi-modal Egocentric Activity Dataset (MEAD) [34] contains 20 distinct life-logging activities grouped into 4 top level types: ambulation, daily activities, office work and exercise. Each activity category has 10 sequences and each clip is exactly 15 seconds. There are a total of 200 videos at 1280x720 at 29.9fps. Audio was sampled at 48 kHz and 16 bits per sample. As no significant content was observed in higher frequencies, we down-sampled it to 24kHz before the feature extraction.

DogCentric Activity Dataset (DogC) [29] is composed of dog activity videos taken from the viewpoint of the dogs. The dataset contains 10 different types of activities performed by dogs. Video resolutions are 320x240. Some of the activities are: playing with a ball, drinking, feeding, looking left/right, petting, and shaking. Unlike the other datasets, the number of videos for each activity is different (i.e. feed and shake have 25 videos while playing with a ball has only 14 videos) which makes this dataset unbalanced.

The performance of the proposed solution was evaluated comparatively for three different egocentric video datasets. Global and local video and audio features were extracted as one feature vector corresponding to each video segment. Therefore, number of samples is equal to the number of videos in datasets.

The average score of 100 test trials were used as the final score. At each test trial, training and test sets are randomly decomposed as 75% and 25% for each activity. For JPL, each activity has 9 training and 3 test videos whereas each activity has 8 training and 2 test videos for MEAD. However, the number of training and test videos varies for DogC since it has different number of samples for each activity.

The performance measures were calculated using various types of metrics such as precision ($P$), recall ($R$), accuracy ($A$) and F1-score ($F$) shown in (6) by considering true positive ($TP$), true negative ($TN$), false positive ($FP$) and false negative ($FN$) scores.

$$P = \frac{TP}{TP + FP} \qquad R = \frac{TP}{TP + FN}$$
$$A = \frac{TP + TN}{TP + TN + FP + FN} \qquad F = \frac{2 * P + R}{P + R} \qquad (6)$$

The Kappa statistic is known to be a discerning statistical tool for assessing the classification accuracy of different classifiers. It was shown that the Kappa statistic is a statistically more sophisticated measure of inter-classifier agreement than the overall accuracy and gives better interclass discrimination than the overall accuracy [35].

Kappa statistic is calculated by using the marginal probabilities of ground truth and predicts labels with their joint probabilities that correspond to the values of confusion matrix. The formulation of Kappa statistic is given below:

$$p_0 = \sum_{i=1}^{C} p_{ii} \qquad p_e = \sum_{i=1}^{C} p_i * p_{\hat{i}} \qquad \hat{\kappa} = \frac{p_0 - p_e}{1 - p_e} \qquad (6)$$

where $C$ is the number of classes, $p_i$ is the probability of i[th] class according to the ground truth, $p_{\hat{i}}$ is the probability of i[th] class according to the prediction, $p_0$ is the observed accuracy and $p_e$ is the sum of the marginal proportions.

Squared Inter-class Confusion (*SIC*) is a new performance evaluation metric we propose to measure whether the resulting confusion matrix concentrates on diagonal or displays a scattered pattern at off-diagonal members. Its formulation is given as:

$$SIC = 1 - \frac{1}{C * 100^2} \sum_{i=1}^{C} \left( \hat{v}_{diag}^{i} - 100 \right)^2 \quad (7)$$

where $\hat{v}_{diag}$ is the diagonal elements of the resulting confusion matrix in terms of percentage and $C$ is the total number of classes. If the resulting confusion matrix is similar to the ideal confusion matrix, the score gets closer to 1. Otherwise, SIC value gets closer to 0.

Implementation was done mainly using MATLAB while OpenCV (3.2.0) was used for optical flow estimation, MATLAB toolbox developed by Dollar et al. [21] was used for *cuboid* extraction and LibSVM (v0.9.20) [36] was used for *SVM* classification. LibSVM library has been modified to support histogram intersection kernels.

We first analyzed the performances of individual features. Then, we analyzed the combination of features to observe the effects of feature types on the recognition scores. The results were tabulated for each dataset individually. Finally, we evaluated the performance of the proposed method when all the features are used and compared with the results in the literature.

## A. JPL

Average F1-scores of single features and the combinations of features on JPL dataset are shown in TABLE I. The results show that SimpleMKL algorithm performs better for optical flow based features while histogram intersection kernels work better for histogram based features (*Log-C* and *cuboid*). On the other hand, *cuboid* feature performance falls behind the global video features.

TABLE I
F1-Score of Single Features and Feature Combinations for JPL

| | | SVM-Poly | SVM-Hist | SimpleMKL | MKBoost |
|---|---|---|---|---|---|
| Single Features | | | | | |
| Global | GOFF | 0.91 | 0.91 | **0.92** | **0.92** |
| Global | VIF | 0.85 | 0.86 | **0.87** | 0.86 |
| Global | Log-C | 0.80 | **0.84** | 0.82 | 0.82 |
| Local | Cuboid | 0.69 | **0.71** | 0.70 | 0.70 |
| Combination of Global Features | | | | | |
| | GOFF + VIF | 0.92 | 0.92 | **0.93** | **0.93** |
| | GOFF + Log-C | 0.90 | 0.87 | 0.92 | **0.93** |
| | VIF + Log-C | 0.84 | **0.86** | 0.85 | 0.85 |
| | GOFF + VIF + Log-C | 0.91 | 0.89 | **0.93** | **0.93** |
| Combination of Global and Local Features | | | | | |
| | Global + Local | 0.92 | 0.92 | **0.93** | **0.93** |

In general, global features have better results compared to the local feature cuboid. Global feature combinations including *GOFF* consistently got the highest scores. This result is expected since *GOFF* is the most discriminative feature according to the individual feature performance results. When all features (global+local) are used, *MKL* and *MKBoost* algorithms work better compared to the others.

The resulting confusion matrices of JPL activities for different methods are shown in Fig. 2 in the order of *hug*, *pet*, *point*, *punch*, *shake*, *throw* and *wave*. According to these matrices, pet and point activities had the worst recognition performances compared to other activities. Additionally, *MKBoost* algorithm achieves more balanced recognition performance scores for all activities.

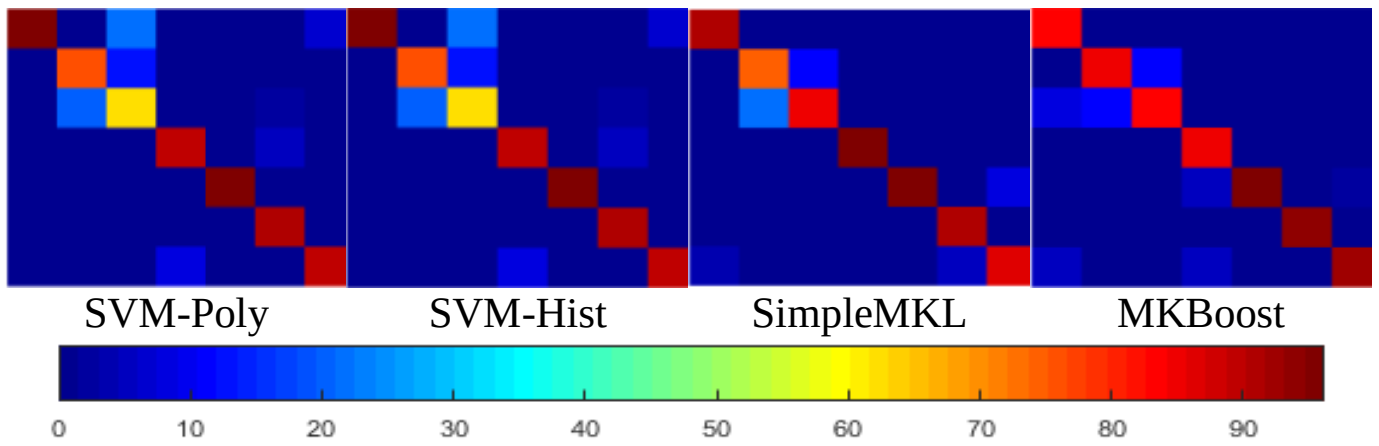

Fig. 2. Confusion matrices of SVM, Histogram Intersection, SimpleMKL and MKBoost learning methods for JPL.

## B. MEAD

MEAD is a more difficult dataset for activity recognition compared to JPL since it has a more diverse set of egocentric activities. Unlike the other datasets, audio features were also used for MEAD in addition to the global and local video features. The results showed that the characteristics of the test results are quite similar to JPL for the individual features. For example, *MKL* based methods produced better results for *GOFF* and *VIF* (TABLE II). On the other hand, Simple MKL and histogram intersection kernel gave similar results for *Log-C* feature while histogram intersection had better classification performance for cuboid. For audio features, the highest score was obtained with SimpleMKL.

TABLE II
F1-Score of Single Features and Feature Combinations for MEAD

| | | SVM-Poly | SVM-Hist | SimpleMKL | MKBoost |
|---|---|---|---|---|---|
| Single Features | | | | | |
| Global | GOFF | 0.55 | 0.55 | 0.56 | **0.57** |
| Global | VIF | 0.26 | 0.30 | **0.31** | 0.28 |
| Global | Log-C | 0.35 | 0.36 | **0.37** | 0.35 |
| Local | Cuboid | 0.27 | **0.33** | 0.29 | 0.30 |
| Audio | Audio | 0.42 | 0.42 | **0.44** | 0.43 |
| Combination of Global Features | | | | | |
| | GOFF + VIF | 0.57 | 0.54 | **0.58** | **0.58** |
| | GOFF + Log-C | **0.60** | 0.56 | 0.59 | **0.60** |
| | VIF + Log-C | 0.42 | 0.43 | **0.44** | 0.42 |
| | GOFF + VIF + Log-C | 0.61 | 0.59 | **0.62** | 0.61 |
| Combination of Global, Local and Audio Features | | | | | |
| | Global + Local | **0.63** | 0.60 | **0.63** | 0.61 |
| | Global + Audio | 0.64 | 0.62 | **0.70** | 0.65 |
| | Local + Audio | 0.45 | **0.53** | 0.50 | **0.53** |
| | Global + Local + Audio | 0.65 | 0.67 | **0.71** | 0.68 |

Similar to JPL, MKL-based learning algorithms performed better for global video features compared to SVM-based approaches. When combinations of global and local video and audio features are used, SimpleMKL outperforms other classifiers by better assigning the weights for multi-modal features.

The resulting confusion matrices for different methods are shown in Fig. 3. Activities are given in the order of *cycling*, *doing push-ups*, *doing sit up*, *drinking*, *eating*, *making phone* *calls*, *organizing files*, *reading*, *riding elevator down*, *riding elevator up*, *riding escalator down*, *riding escalator up*, *running*, *sitting*, *texting*, *walking*, *walking downstairs*, *walking upstairs*, *working at PC* and *writing sentences*. The results showed that the recognition performance is not good when the motion is low (i.e. *reading*, *organizing files* or *texting*) while the scores get better when the motion is high (i.e. *walking*, *doing push-ups*, or *walking downstairs*).

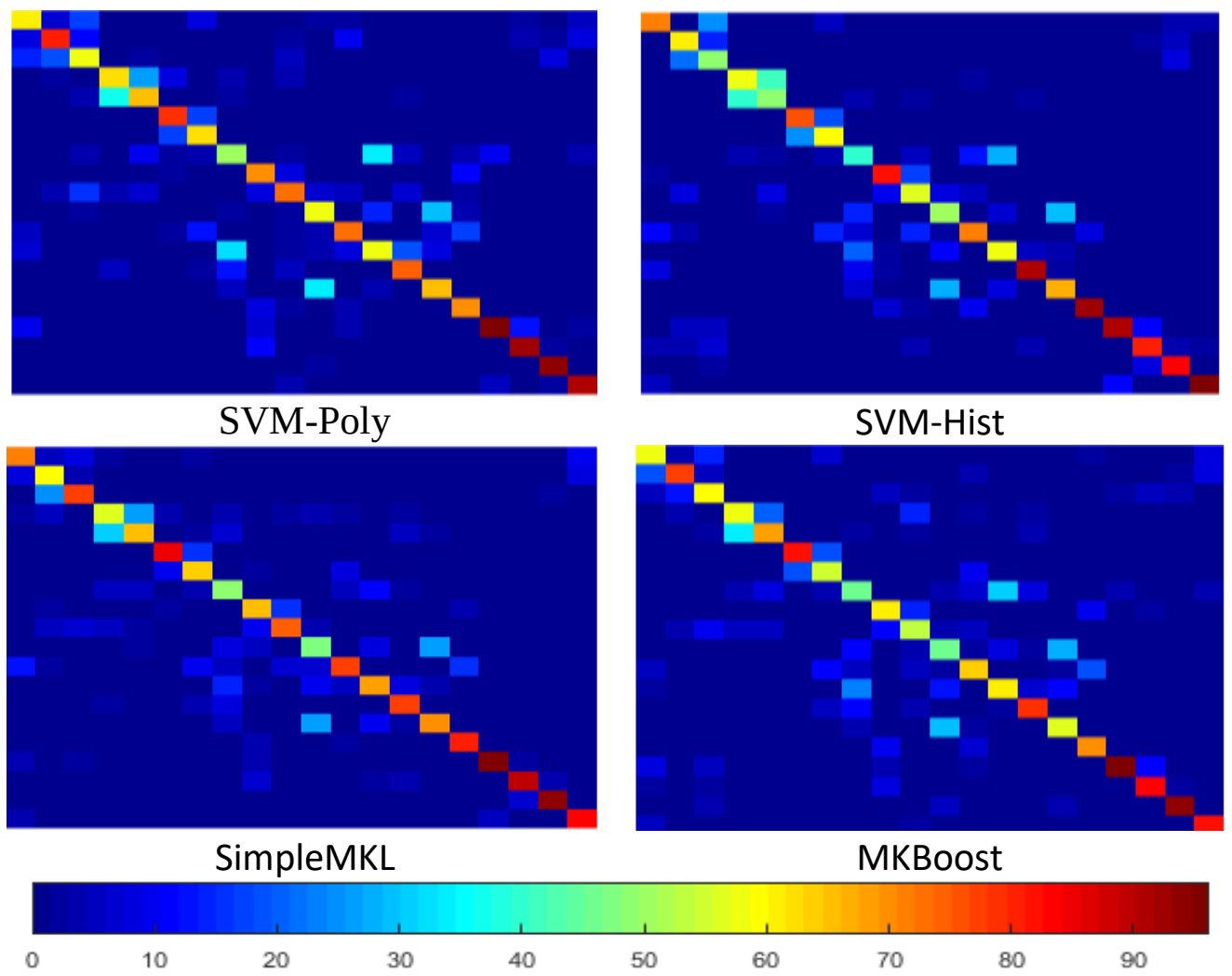

Fig. 3. Confusion matrices of polynomial SVM, histogram intersection and SimpleMKL algorithms for MEAD.

### C. DogC

Unlike the other datasets, DogC is an unbalanced dataset since it has a varying number of videos for each activity. TABLE III shows average F1-scores of individual feature performances, and combinations of features on DogC dataset. Similar to MEAD and JPL, SimpleMKL had better scores for GOFF and VIF. Histogram intersection kernel performed better for Log-C, but cuboid feature performed better with the polynomial kernel. Similar to the previous results, SimpleMKL performs better for the combination of global video features and also it has the best results when all the features are used.

TABLE III
F1-Score of Single Features and Feature Combinations for DogC

| | | SVM-Poly | SVM-Hist | SimpleMKL | MKBoost |
|---|---|---|---|---|---|
| Single Features | | | | | |
| Global | GOFF | 0.56 | 0.59 | **0.61** | 0.59 |
| Global | VIF | 0.42 | 0.46 | **0.47** | **0.47** |
| Global | Log-C | 0.47 | 0.51 | 0.49 | **0.53** |
| Local | Cuboid | **0.43** | 0.36 | 0.41 | 0.41 |
| Combination of Global Features | | | | | |
| GOFF + VIF | | 0.60 | 0.61 | **0.63** | **0.63** |
| GOFF + Log-C | | 0.59 | **0.63** | **0.63** | 0.62 |
| VIF + Log-C | | 0.53 | 0.52 | 0.55 | **0.56** |
| GOFF + VIF + Log-C | | 0.62 | **0.63** | **0.63** | **0.63** |
| Combination of Global and Local Features | | | | | |
| Global + Local | | 0.64 | 0.62 | **0.65** | 0.63 |

The resulting confusion matrices of DogC activities for different learning methods are shown in Fig. 4. Activities are given in the order of *playing with a ball*, *waiting for a car to pass by*, *drinking water*, *feeding*, *looking left*, *looking right*, *petting*, *shaking dog's body*, *sniffing*, *walking*. "Looking left" and "looking right" activities have the lowest classification accuracies since they are confused with one another. These two activities have very similar characteristics apart from having different motion directions.

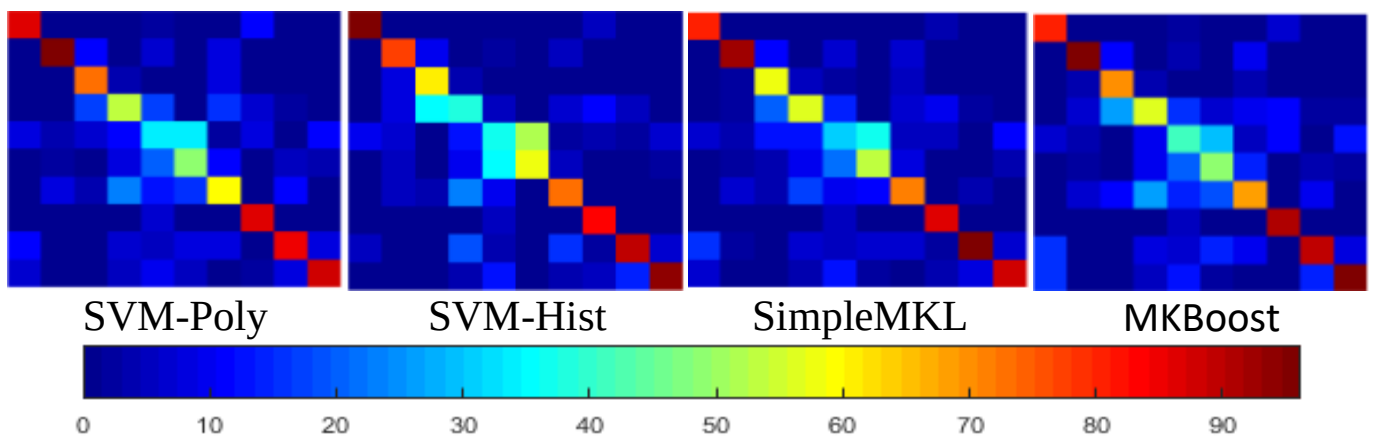

Fig. 4. Confusion matrices of SVM, Histogram Intersection, SimpleMKL and MKBoost learning methods for DogC.

### D. Comparative Results

In this section, the performance results are given in TABLE IV when all features are used. In this table, average accuracy ($\mathcal{A}$), precision ($\mathcal{P}$), recall ($\mathcal{R}$), Kappa value ($\kappa$) and F1-scores ($\mathcal{F}$) are shown for three egocentric datasets.

MKL methods' classification results were better when compared to SVM approaches for most of the evaluation metrics – particularly pronounced after adding audio information in addition to video features for MEAD dataset.

For JPL, the proposed method outperforms the other similar works [9] [10]. In this work, motion (GOFF) and inertia (VIF) based features are the same with [9] except that we performed the motion estimation algorithm by using Farneback while Horn-Schunk motion estimation algorithm was used in [9]. On the other hand, in [10], a different approach called structural learning is also used. Even though the overall accuracies were close to each other, MKL methods had higher accuracy values.

For MEAD, Song et al. [37] had an accuracy result of 0.84 using 19 additional sensor data including accelerometer, gravity, gyroscope, linear acceleration, magnetic field and rotation vector.

TABLE IV
Comparative Performances of the Proposed Method

| Dataset | Methods | $\mathcal{A}$ | $\mathcal{P}$ | $\mathcal{R}$ | $\kappa$ | $SIC$ | $\mathcal{F}$ |
|---|---|---|---|---|---|---|---|
| **JPL** | SVM | 0.91 | 0.92 | 0.91 | 0.89 | **0.99** | 0.91 |
| | DC-Int | 0.87 | 0.90 | 0.87 | 0.87 | 0.97 | 0.89 |
| | SimpleMKL | **0.92** | 0.93 | **0.92** | **0.91** | **0.99** | **0.93** |
| | MKBoost | **0.92** | **0.94** | **0.92** | **0.91** | **0.99** | **0.93** |
| | Abebe et al. [9] | - | 0.87 | 0.85 | - | - | 0.86 |
| | Ryoo&Matthies [10] | 0.90 | - | - | - | - | - |
| | Ozkan et al. [16] | 0.87 | - | - | - | - | - |
| | Sudhakaran & Oswald [12] | 0.91 | - | - | - | - | - |
| **MEAD** | SVM | 0.65 | 0.65 | 0.65 | 0.63 | 0.86 | 0.65 |
| | DC-Int | 0.66 | 0.67 | 0.66 | 0.66 | 0.87 | 0.67 |
| | SimpleMKL | **0.70** | **0.71** | **0.70** | **0.70** | **0.90** | **0.71** |
| | MKBoost | 0.68 | 0.69 | 0.68 | 0.66 | 0.88 | 0.68 |
| **DogC** | SVM | 0.63 | 0.65 | 0.63 | 0.58 | 0.83 | 0.64 |
| | DC-Int | 0.60 | 0.64 | 0.60 | 0.59 | 0.83 | 0.62 |
| | SimpleMKL | 0.64 | **0.66** | **0.65** | **0.61** | **0.84** | **0.65** |
| | MKBoost | 0.61 | 0.64 | 0.63 | 0.57 | 0.82 | 0.63 |
| | Abebe et al. [9] | - | 0.62 | 0.59 | - | - | 0.61 |

| | | | | | | |
|---|---|---|---|---|---|---|
| Iwashita et al. [29] | 0.61 | - | - | - | - | - |
| Ozkan et al. [16] | **0.65** | - | - | - | - | - |

DogC is a more challenging dataset as the ego-motion in the videos are quite large and it is unbalanced (varying number of videos for different activities). Abebe et al. [9] applied the same methodology used for JPL. Iwashita et al. [29] used global (dense optical flow and local binary pattern) and local (normalized pixel values, HOG and HOF) motion descriptors and combined them with a modified histogram intersection kernel. The results showed that SimpleMKL approach has better performance values compared to the other methods.

## IV. Discussion

The results showed that the performance of the proposed framework was mostly better compared to the state-of-the art methods for the three egocentric datasets. Even though the single feature performances vary for different learning algorithms, MKL approaches (SimpleMKL and MKBoost) outperform the others when global/local video and audio features are combined. That makes MKL a prominent method for the fusion of features and confirms its ability to fuse features efficiently.

Even though only three types of sensor data (video, audio and video-based inertia) were used in this work, other types of sensor data (such as eye-tracking sensors, magnetometers, proximity sensors, temperature sensors) can also be used with MKL approach since MKL provides easy integration of new features. New features are considered as new channels of information to be adaptively learned by the base learners. Each feature is assigned a weight with respect to its classification performance. By this way, feature selection and model training are done concurrently.

Another important point is that the use of multiple modalities (i.e., video and audio) improved the activity recognition performance which also means multi-modal features have complementary information coming from different domains. Additionally, the experimental results showed that MKL is the most effective solution when using multi-modal features compared to other classifiers. For example, combining global or local video features with audio for MEAD dataset provided a significant improvement for MKL compared to other classifiers.

In order to understand the kernel and feature selection of MKL, histogram of the selected kernels (Fig. 5-(a)) and features (Fig. 5-(b)) for the average over all 100 trials of MKBoost algorithm were plotted.

Fig. 5-(a) shows that linear and polynomial were the most dominantly selected base kernels for JPL and MEAD datasets. For DogC, linear, RBF and histogram intersection kernels were selected the most followed by polynomial kernels.

In Fig. 5-(b), the features were assumed to be selected if they belong to the selected feature combination set. For example, GOFF was marked as selected when any feature combination that includes GOFF (e.g., GOFF+VIF, GOFF+VIF+Log-C) was selected. It was observed that optical flow-based GOFF was the most discriminative feature among others as its selection rates were high for all datasets. Unlike the other datasets, cuboid feature was the most selected feature for DogC. However, the selection rates of all features were not significantly different from each other. It could also be said that audio feature had significant contribution to the recognition performance since it was the second most selected feature for MEAD. The features in the order of their selection frequency for different datasets were: MEAD: GOFF > Audio > Cuboid > LogC > VIF; JPL: GOFF > VIF > LogC > Cuboid, DogC: Cuboid > GOFF > VIF > LogC. The difference in ordering for different datasets shows that the classification was able to adapt to the characteristics of different types of data. While the global features give the best information for classification of human activities, local feature was more important than the global features for DogC dataset due to the more hectic nature of dog motions and when audio information is available it also becomes an important element.

More detailed analyses for kernel and feature selection were performed to analyze which feature combinations were mostly used with which base kernels. For that purpose, the histogram of the selected feature combinations and corresponding base kernel selection were extracted for each dataset (Fig. 6). The IDs for feature combinations are given in TABLE V.

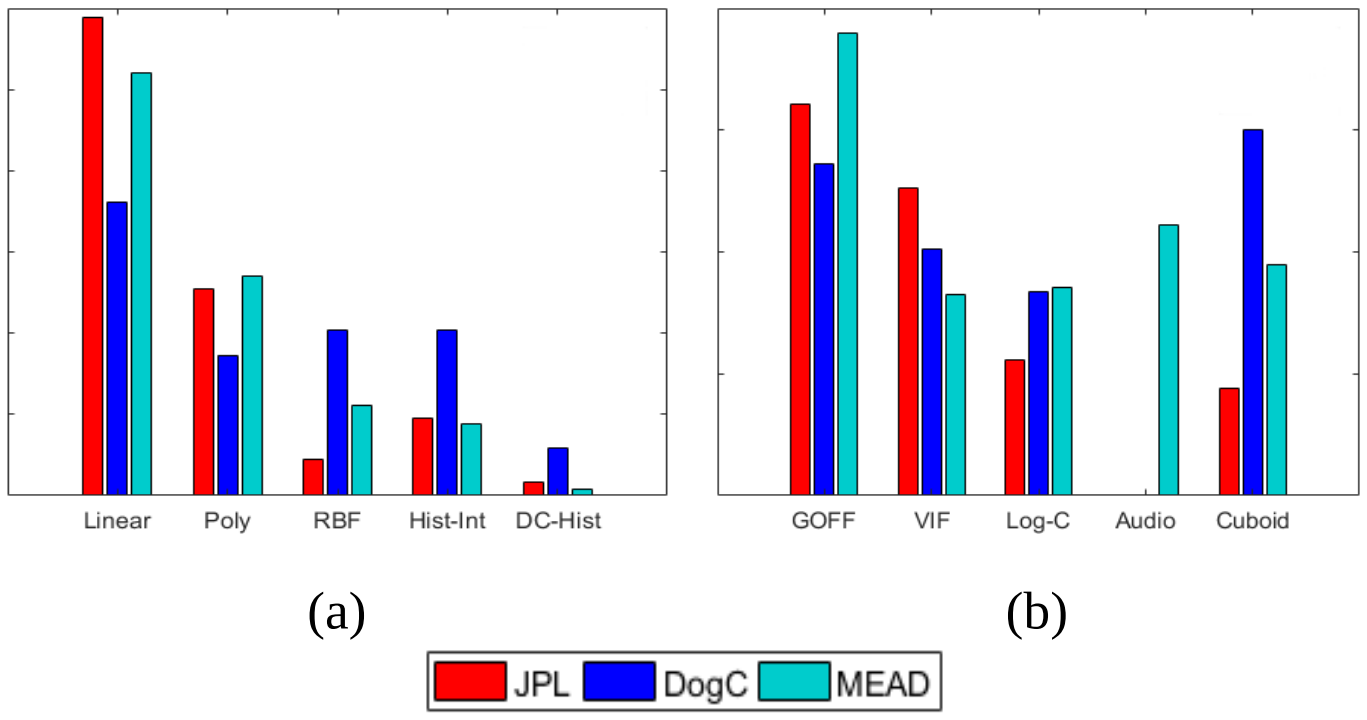

Fig. 5. The number of selected (left) base kernel and (right) feature histograms.

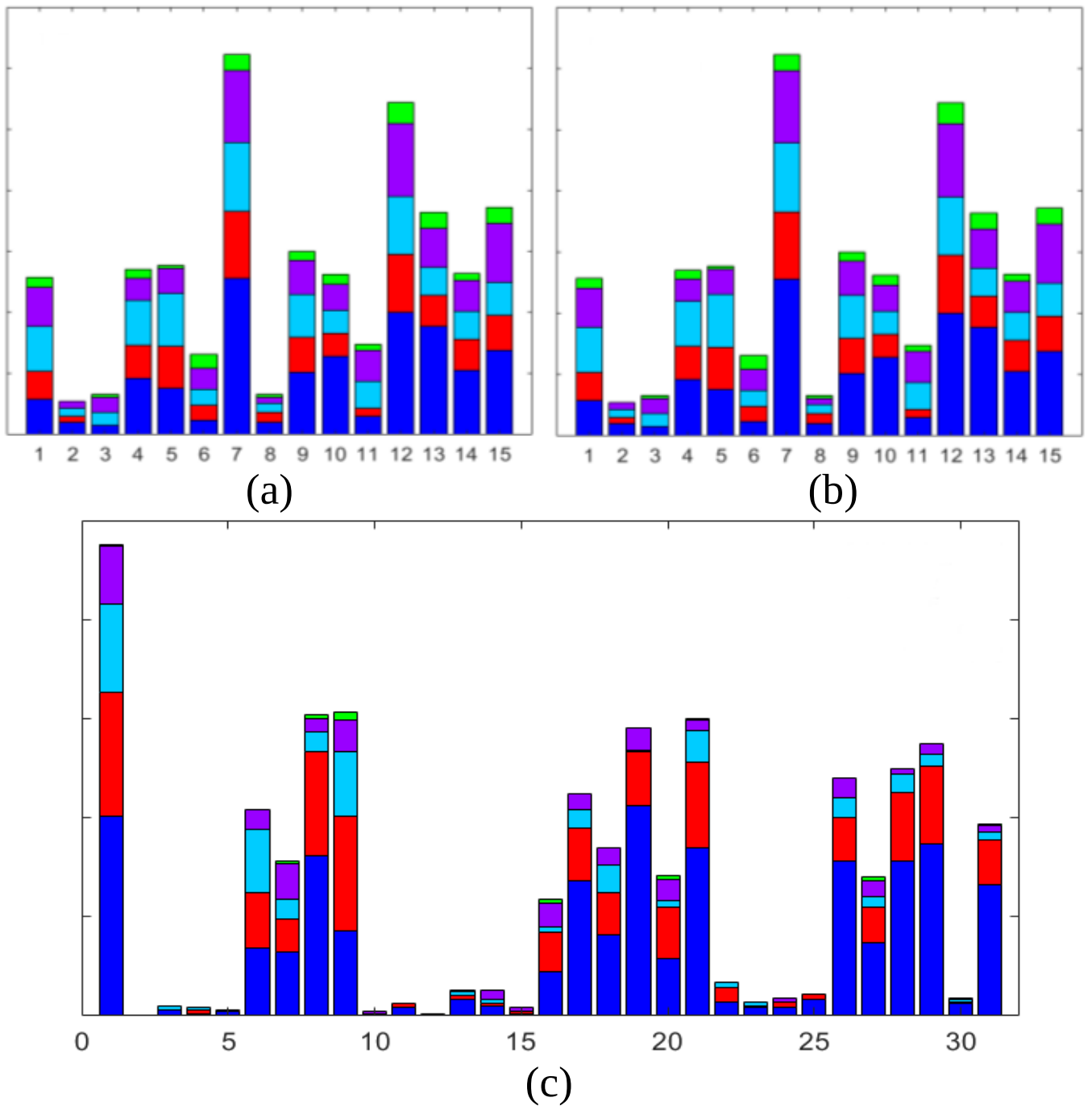

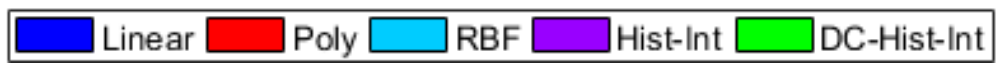

Fig. 6. The number of selected feature combinations for JPL(top-left), DogC(top-right) and MEAD(bottom).

Although the feature combination IDs are the same for JPL and DogC, they are different for MEAD since it has an additional audio feature. TABLE V shows the feature combination IDs and their feature compositions. The related feature combination includes the feature if it is colored as green.

Fig. 6-(a) shows that GOFF, VIF and GOFF+VIF were the most selected features for JPL. For DogC, cuboid feature was selected as a single feature and was included in the most selected feature combinations according to Fig. 6-(b). On the other hand, GOFF was selected nearly for all of the selected feature combinations for MEAD showing that optical flow-based spatio-temporal feature can model the egocentric activities better than other features. Another point that needs to be emphasized is that although audio was the second most selected feature (Fig. 5-(b)) for MEAD, it was selected as combinatory information with other features instead of a single feature.

Even if MKL approaches got satisfactory results for egocentric activity recognition problem, there are some disadvantages of using them. Firstly, it is hard to configure pre-defined basis kernels. If the basis kernels are not selected properly, MKL approaches may not find the optimal solution. In this work, the same base kernel set was used for MKBoost algorithm. Besides that, computation time for training becomes longer for MKBoost algorithm, because feature combinations are tested for each trial and for all basis kernels. The time needed for one test trial is directly proportional to all feature combinations, number of basis kernels and number of trials.

TABLE V
FEATURE COMBINATION IDS AND FEATURE COMPOSITIONS. FEATURE NAMES ARE ABBREVIATED AS FOLLOWS: GOFF:G, VIF:V, LOG-C:L, CUBOID:C AND AUDIO:A

| **JPL & DogC** | | | | | **MEAD** | | | | | | | | | | | |
|---|---|---|---|---|---|---|---|---|---|---|---|---|---|---|---|---|
| ID | G | V | L | C | ID | G | V | L | A | C | ID | G | V | L | A | C |
| 1 | ■ | | | | 1 | ■ | | | | | 16 | ■ | ■ | ■ | | |
| 2 | | ■ | | | 2 | | ■ | | | | 17 | ■ | ■ | | ■ | |
| 3 | ■ | ■ | ■ | ■ | 3 | | | ■ | | | 18 | ■ | ■ | | | ■ |
| 4 | | | | ■ | 4 | | | | ■ | | 19 | ■ | | ■ | ■ | |
| 5 | ■ | ■ | | | 5 | | | | | ■ | 20 | ■ | | ■ | | ■ |
| 6 | ■ | | ■ | | 6 | ■ | ■ | | | | 21 | ■ | | | ■ | ■ |
| 7 | ■ | | | ■ | 7 | ■ | | ■ | | | 22 | | ■ | ■ | ■ | |
| 8 | | ■ | ■ | | 8 | ■ | | | ■ | | 23 | | ■ | ■ | | ■ |
| 9 | | ■ | | ■ | 9 | ■ | | | | ■ | 24 | | ■ | | ■ | ■ |
| 10 | | | ■ | ■ | 10 | | ■ | ■ | | | 25 | | | ■ | ■ | ■ |
| 11 | ■ | ■ | ■ | | 11 | | ■ | | ■ | | 26 | ■ | ■ | ■ | ■ | |
| 12 | ■ | ■ | | ■ | 12 | | ■ | | | ■ | 27 | ■ | ■ | ■ | | ■ |
| 13 | ■ | | ■ | ■ | 13 | | | ■ | ■ | | 28 | ■ | ■ | | ■ | ■ |
| 14 | | ■ | ■ | ■ | 14 | | | ■ | | ■ | 29 | ■ | | ■ | ■ | ■ |
| 15 | ■ | ■ | ■ | ■ | 15 | | | | ■ | ■ | 30 | | ■ | ■ | ■ | ■ |
| | | | | | | | | | | | 31 | ■ | ■ | ■ | ■ | ■ |

## V. CONCLUSION

In this work, we proposed a new framework for egocentric activity recognition problem based on audio-visual features combined with multi-kernel learning classification. It was shown that using audio features in addition to global and local video features improves the recognition performance. Additionally, MKL approach performed well for the fusion of features extracted from different domains. The proposed solution was tested on three different egocentric video datasets. The proposed method had mostly better results compared to state-of-the-art methods. As a result, it can be said that audio-visual features can be fused with MKL approaches to effectively recognize egocentric activities.

The future work includes adding more informative audio-visual features into the framework. By this way, the features may be more informative about the egocentric activities in videos. Another improvement may be realized by using alternative MKL algorithms for kernel selection and optimization of the recognition problem.

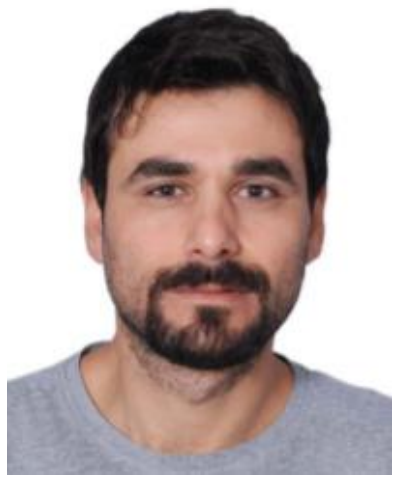

**Mehmet Ali Arabacı** received his B.Sc. and M.Sc. degrees in Electrical & Electronics Engineering from Dokuz Eylül University, Turkey in 2005 and 2008 respectively. He is currently pursuing his Ph.D. degree at Department of Information Systems in METU. He is working as a Senior Researcher in The Scientific and Technological Research Council of Turkey (TUBITAK) since 2011. His current research interests include egocentric video analysis and multi-kernel learning.

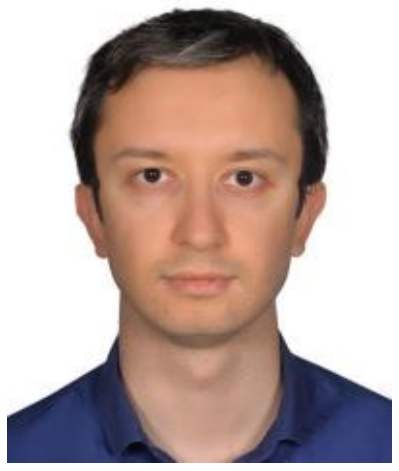

**Fatih Ozkan** received his B.Sc. degree in Computer Engineering from Cukurova University, Turkey in 2012 and the M.Sc. degree in the Department of Information Systems, METU where he is currently working toward the Ph.D. degree. He is working as a Researcher in The Scientific and Technological Research Council of Turkey (TUBITAK) since 2015. His research interests include computer vision and machine learning.

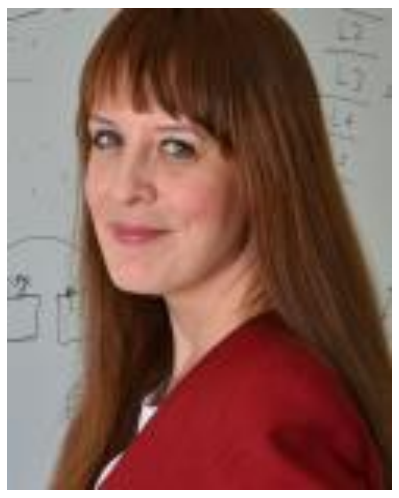

**Elif Surer** received her Ph.D in Bioengineering in 2011 from the University of Bologna. She received her M.S. and B.S. degrees in Computer Engineering from Boğaziçi University in 2007 and 2005, respectively. She is currently working as an Assistant Professor at the METU Graduate School of Informatics' Multimedia Informatics program. She is funded by the H2020 project eNOTICE as METU PI as of September 2017. Her research interests are serious games, virtual/augmented reality, human and canine movement analysis, machine learning and computer vision.

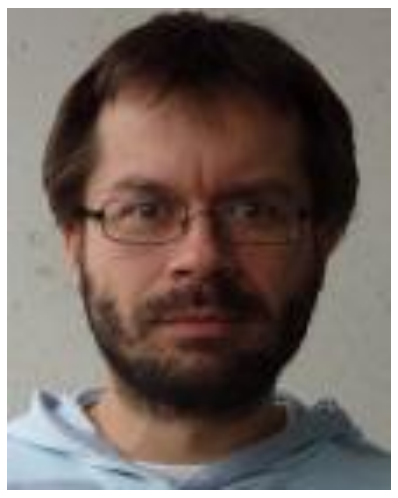

**Peter Jančovič** received the B.Sc. and M.Sc. degrees in information technology from the Slovak University of Technology, Bratislava, Slovakia, in 1997 and 1999, respectively, and the Ph.D. degree in the field of noise robust speech recognition from the School of Computer Science, Queen's University Belfast, Northern Ireland, U.K., in 2002. He is currently a Lecturer with the Department of Electronic, Electrical and Computer Engineering, University of Birmingham, U.K.

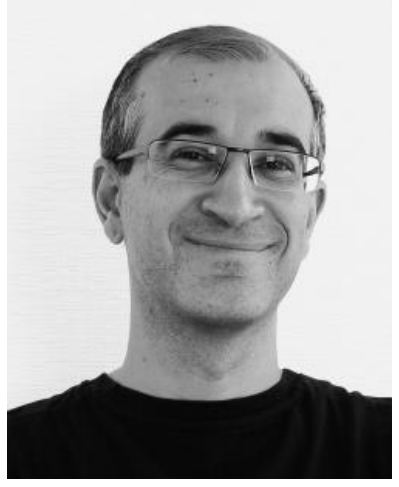

**Alptekin Temizel** Assoc. Prof., Graduate School of Informatics, Middle East Technical University (METU) B.Sc., Electrical and Electronics Engineering, METU (1999); Ph.D. Centre of Vision, Speech and Signal Processing, University of Surrey (2006). He is the principal investigator of GPU Education and Research Centre and a Deep Learning Institute Certified Instructor and University Ambassador. He collaborated with and advised several companies on video analytics, video surveillance systems/algorithms and GPU computing. He was a visiting researcher at Microsoft MLDC-Lisbon in the summers of 2014 and 2015. He was on sabbatical leave at University of Birmingham, UK in 2016-2017. His main research interests are video surveillance, computer vision, machine learning, deep learning, GPU computing.